\documentclass[11pt,letterpaper]{article}
\usepackage[letterpaper]{geometry}
\usepackage{acl2012}
\usepackage{times}
\usepackage{latexsym}
\usepackage{amsmath,mathtools}
\usepackage{amssymb}
\usepackage{graphicx}
\usepackage{ifthen}
\usepackage[utf8]{inputenc}
\usepackage{url}
\usepackage{tikz}
\usepackage{tikz-cd}
\usepackage{subcaption}
\usepackage{pgfplots,capt-of}
\usepackage{multirow}
\usepackage{array,booktabs}
\usepackage{enumitem}
\usepackage{wrapfig}
\usepackage{stmaryrd}
\usepackage{adjustbox}
\usepackage{filecontents}
\usepackage{pgfplotstable}

\pgfplotsset{compat=newest}
\usetikzlibrary{matrix, calc, arrows, shapes.geometric,shapes.callouts,shapes,positioning,shapes.misc,pgfplots.statistics}

\makeatletter
\newcommand{\@BIBLABEL}{\@emptybiblabel}
\newcommand{\@emptybiblabel}[1]{}
\makeatother
\usepackage[hidelinks]{hyperref}

\title{Planning, Inference and Pragmatics in Sequential Language Games}

\author{Fereshte Khani \\
  Stanford University \\
  {\small \tt fereshte@stanford.edu} \\\And
  Noah D. Goodman \\
  Stanford University \\
  {\small \tt ngoodman@stanford.edu} \\\And
  Percy Liang \\
  Stanford University\\
  {\small \tt pliang@cs.stanford.edu} \\}

\date{}

\newcommand\sz{\ensuremath{\mathcal{z}}}

\newcommand\R{\ensuremath{\mathbb{R}}} 
\newcommand\eqdef{\ensuremath{\stackrel{\rm def}{=}}} 
\newcommand{\1}{\mathbb{I}} 

\newcommand\refsec[1]{Section~\ref{sec:#1}}

\newcommand\reffig[1]{Figure~\ref{fig:#1}}

\newcommand\reftab[1]{Table~\ref{tab:#1}}

\ifthenelse{\isundefined{\definition}}{\newtheorem{definition}{Definition}}{}
\ifthenelse{\isundefined{\assumption}}{\newtheorem{assumption}{Assumption}}{}
\ifthenelse{\isundefined{\hypothesis}}{\newtheorem{hypothesis}{Hypothesis}}{}
\ifthenelse{\isundefined{\proposition}}{\newtheorem{proposition}{Proposition}}{}
\ifthenelse{\isundefined{\theorem}}{\newtheorem{theorem}{Theorem}}{}
\ifthenelse{\isundefined{\lemma}}{\newtheorem{lemma}{Lemma}}{}
\ifthenelse{\isundefined{\corollary}}{\newtheorem{corollary}{Corollary}}{}
\ifthenelse{\isundefined{\alg}}{\newtheorem{alg}{Algorithm}}{}
\ifthenelse{\isundefined{\example}}{\newtheorem{example}{Example}}{}

\newcommand\citep\cite
\newcommand\citet\newcite

\newcommand\den[1]{\llbracket #1 \rrbracket}

\newcommand\pl[1]{\textcolor{red}{[PL: #1]}}
\newcommand\fk[1]{\textcolor{cyan}{[FK: #1]}}

\newcommand\Pchar{\ensuremath{P_\text{letter}}}
\newcommand\Pnum{\ensuremath{P_\text{digit}}}

\newcommand\spa[1]{\texttt{#1}}   

\newcommand\gameName{InfoJigsaw}
\newcommand\modelName{PIP}
\newcommand\noPrag{\ensuremath{\text{PIP}_{\text{-prag}}}}
\newcommand\noPlan{\ensuremath{\text{PIP}_{\text{-plan}}}}
\newcommand\noInfer{\ensuremath{\text{PIP}_{\text{-infer}}}}

\newcommand\refeqn[1]{(Eqn. \ref{eqn:#1})}

\newcommand\bm[3]{\ensuremath{\left[\begin{smallmatrix} #1 \\ #2\\ #3 \end{smallmatrix}\right]}}

\tikzset{
	objectC/.style={scale=\sc, circle, fill=yellow, draw=black, minimum size=0.9cm, inner sep=0},
	objectS/.style={scale=\sc, regular polygon,regular polygon sides=4, fill=yellow, draw=black, minimum size=1.2cm, inner sep=0},
	objectD/.style={scale=\sc, diamond, fill=yellow, draw=black, minimum size=1.1cm, inner sep=0},
	goal/.style={black},
	message/.style={black, anchor=west,inner sep=0},
	cloudStyle/.style={anchor=west,scale=\sc, cloud, cloud puffs=26,draw, inner sep=0em, cloud ignores aspect, align=left,text width=8.5em},
}

\begin{document}

\maketitle

\begin{abstract}
We study sequential language games in which two players, each
with private information, communicate to achieve a common goal.
In such games, a successful player must
(i) infer the partner's private information from the partner's messages,
(ii) generate messages that are most likely to help with the goal, and
(iii) reason pragmatically about the partner's strategy.
We propose a model that captures all three characteristics
and demonstrate their importance in capturing human behavior
on a new goal-oriented dataset we collected using crowdsourcing.

\end{abstract}
\section{Introduction}

Human communication is extraordinarily rich.
People routinely choose what to say based on their goals (planning), figure out the state of the world based on what others say (inference), all while taking into account that others are strategizing agents too (pragmatics).
All three aspects have been studied in both the linguistics and AI communities.
For planning, Markov Decision Processes and their extensions
can be used to compute utility-maximizing actions via forward-looking recurrences (e.g., \citet{vogel2013emergence}).
For inference, model-theoretic semantics \citep{montague73ptq}
provides a mechanism for utterances to constrain possible worlds,
and this has been implemented recently in semantic parsing \citep{matuszek2012grounded,krishnamurthy2013jointly}.
Finally, for pragmatics, the cooperative principle of \citet{grice75maxims}
can be realized by models in which a speaker simulates a listener---e.g., \citet{franke2009signal} and \citet{frank2012pragmatics}.

\begin{figure}[t]
\centering
\def\sc{0.6}
\def\scc{0.3}
\def\sz{2.8}
\def\di{0.01}

     \centering
        \begin{tikzpicture}[scale=\scc]
%
%
%
%
%
%
%

			\node[goal]  at (\sz,3*\sz) {Find B2};
			\node[goal]  at (3*\sz,3*\sz) {Find B2};
			
			\node[objectC,fill=cyan!20] at (\sz,0) 		{B ?};
			\node[objectS,fill=cyan!20] at (\sz,\sz) 	{B ?};
			\node[objectC,fill=cyan!20] at (\sz,2*\sz) 	{C ?};

			\node at (\sz,-\sz) {\small \Pchar{} view };
			
			\draw[dashed] (2*\sz,-\sz) -- (2*\sz,3*\sz);
			
			\node[objectC,fill=cyan!20] at (3*\sz,0) 		{? 2};
			\node[objectS,fill=cyan!20] at (3*\sz,\sz) 	{? 3};
			\node[objectC,fill=cyan!20] at (3*\sz,2*\sz) 	{? 2};

			\node at (3*\sz,-\sz) {\small \Pnum{} view };
			
			\draw (-0.5*\sz,-1.5*\sz) -- (4.5*\sz,-1.5*\sz);
			
			\node[message] (m1) at (0,-2*\sz)	{\Pchar: \small\spa{square}};
			\node[message] (m2) at (0,-2.7*\sz)	{\Pnum: \small\spa{circle}};
			\node[message] (m3) at (0,-3.4*\sz)	{\Pchar: \small\spa{click (1,3)}};

\node [cloudStyle] (c1) at (-11,4) {{\bf Planning:} Let me first try \spa{\small square}, which is just one possibility.};
\node [,cloudStyle] (c2) at (-11,-1.8) {{\bf Inference:} The square's letter must be B.};
\node [cloudStyle,text width=7.5em] (c3) at (-11,-8) {{\bf Pragmatics:} The square's digit cannot be 2.};

\draw[densely dotted] (m1.west) -- (c1.south east);
\draw[densely dotted] (m2.west) -- (c2.south east);
\draw[densely dotted] (m3.west) -- (c3.south east);

\draw (-0.5*\sz,-3.8*\sz) -- (4.5*\sz,-3.8*\sz);
			
			\end{tikzpicture}

\caption{\label{fig:inference}%
A game of \gameName{} played by two human players.
One of the players (\Pchar) only sees the letters, while the other one (\Pnum) only sees the digits.
Their goal is to identify the goal object, B2, by exchanging a few words.
The clouds show the hypothesized role of planning, inference, and pragmatics in the players' choice of utterances.
In this game, the bottom object is the goal (position (1, 3)).
}

\end{figure}
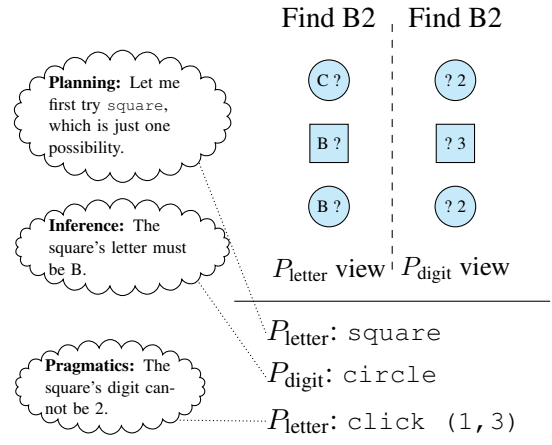

There have been a few previous efforts in the language games literature to combine the three aspects.
\citet{hawkins2015you} proposed a model of communication between a questioner and an answerer based on only one round of question answering.
\citet{vogel2013implicatures} proposed a model of two agents playing a restricted version of the game from the Cards Corpus \cite{potts2012cards},
where the agents only communicate once.\footnote{Specifically, two agents must both co-locate with a specific card. The agent which finds the card sooner shares the card location information with the other agent.}
In this work, we seek to capture all three aspects in a single, unified framework which allows for multiple rounds of communication.

Specifically, we study human communication in a \emph{sequential language game}
in which two players, each with private knowledge, try to achieve a common goal by talking.
We created a particular sequential language game called \gameName{} (\reffig{inference}).
In \gameName{}, there is a set of objects with public properties (shape, color, position) and private properties (digit, letter).
One player (\Pchar) can only see the letters, while the other player (\Pnum) can only see the digits.
The two players wish to identify the goal object, which is uniquely defined by a letter and digit.
To do this, the players take turns talking;
to encourage strategic language, we allow at most two English words at a time.
At any point, a player can end the game by choosing an object.

Even in this relatively constrained game,
we can see the three aspects of communication at work.
As \reffig{inference} shows, in the first turn, since \Pchar{} knows that the game is multi-turn, she simply says \spa{square}, for if the other player does not click on the square, she can try the bottom circle in the next turn ({\bf planning}).
In the second turn, \Pnum{} infers from \spa{square} that the square's letter is probably B ({\bf inference}).
As the digit on the square is not a $2$, she says \spa{circle}.
Finally, \Pchar{} infers that digits of circles are $2$, and in addition she infers from \spa{circle} that the digit on the square is not a $2$ as otherwise, \Pnum{} would have clicked on it ({\bf pragmatics}).
Therefore, she correctly clicks on (1,3).

In this paper, we propose a model that captures planning, inference, and pragmatics for sequential language games, which we call \modelName{}.
Planning recurrences look forward, 
inference recurrences look back,
and pragmatics recurrences look to simpler interlocutors' model.
The principal challenge is to integrate all three types in a coherent way;
we present a ``two-dimensional'' system of recurrences to capture this.
Our recurrences bottom out in a very simple literal semantics (e.g., context-independent meaning of \spa{circle}),
and we rely on the structure of recurrences to endow words with their rich context-dependent meaning.
As a result, our model is very parsimonious and only has four (hyper)parameters.

As our interest is in modeling human communication in sequential language games,
we evaluate \modelName{} on its ability to predict how humans play \gameName{}.\footnote{One could in principle solve for an optimal communication strategy for \gameName{},
but this would likely result in a solution far from human communication.}
We paired up workers on Amazon Mechanical Turk to play \gameName{},
and collected a total of 1680 games.
Our findings are as follows:
(i) \modelName{} obtains higher log-likelihood than a baseline that chooses actions to convey maximum information in each round;
(ii) \modelName{} obtains higher log-likelihood than ablations that remove the pragmatic or the planning components, supporting their importance in communication;
(iii) \modelName{} is better than an ablation with a truncated inference component that forgets the distant past
only for longer games, but worse for shorter games.
The overall conclusion is that by combining a very simple, context-independent literal semantics
with an explicit model of planning, inference, and pragmatics,
PIP obtains rich context-dependent meanings that correlate with human behavior.

\section{Sequential Language Games}
In a sequential language game, there are two players who have a shared world state $w$.
In addition, each player $j \in \{+1,-1\}$ has a private state $s_j$.
At each time step $t = 1, 2, \dots$,
the active player $j(t) = 2(t \text{ mod } 2) - 1$ (which alternates)
chooses an action (including speaking) $a_t$ based on its policy $\pi_{j(t)}(a_t \mid w, s_{j(t)}, a_{1:t-1})$.
Importantly that player $j(t)$ can see her own private state $s_{j(t)}$, but not the partner's $s_{-j(t)}$.
At the end of the game (defined by a terminating action),
both players receive utility $U(w, s_{+1}, s_{-1}, a_{1:t}) \in \mathbb{R}$.
The utility consists of a penalty if players did not reach the goal and a reward if they reached the goal along with a penalty for each action.
Because the players have a common utility function that depends on private information,
they must communicate the part of their private information relevant for maximizing utility.
In order to simplify notation, we use $j$ to represent $j(t)$ in the rest of the paper.

\paragraph{\gameName.}
\begin{figure*}
\centering
\begin{subfigure}[b]{0.47\textwidth}
\includegraphics[height=115pt]{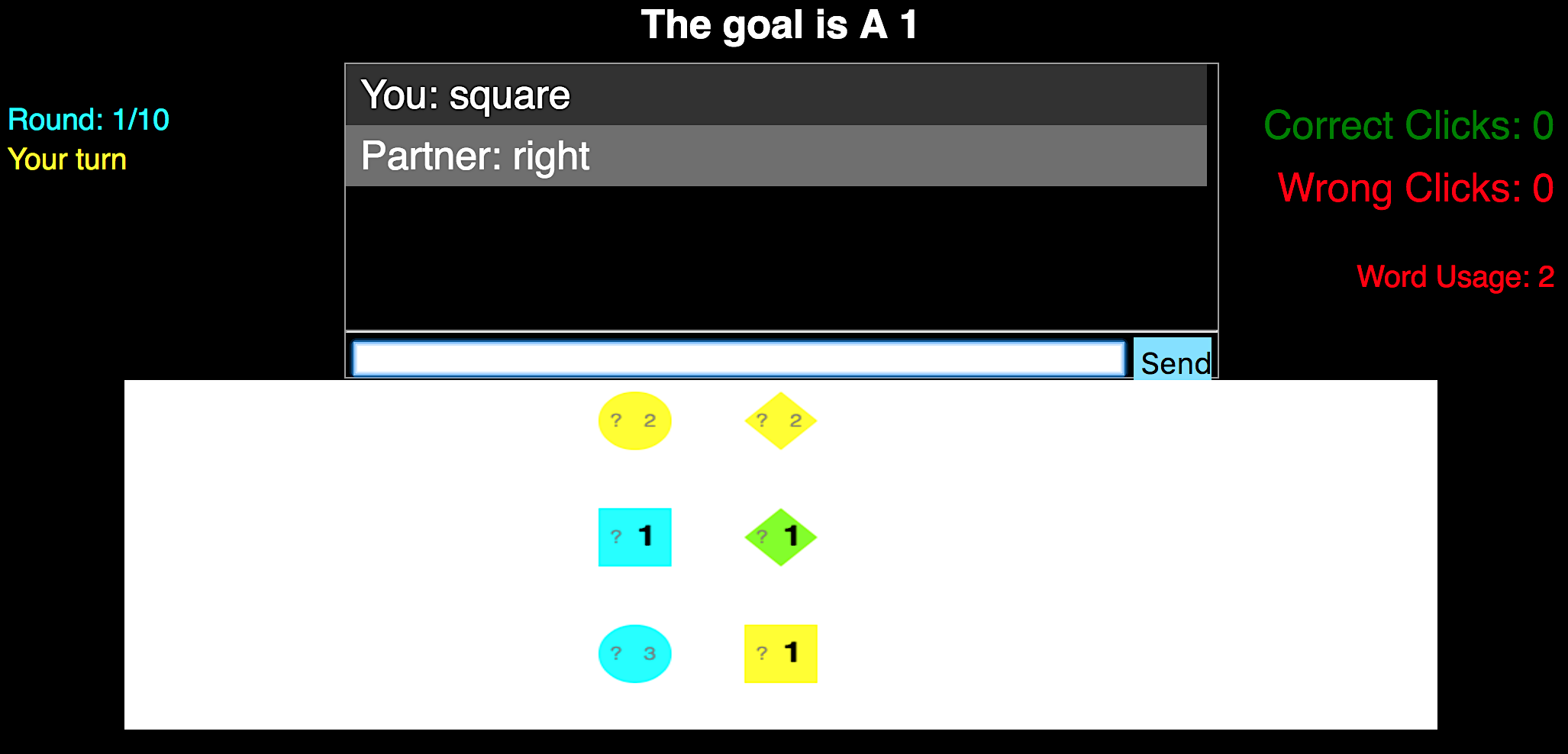}
\caption{\label{fig:pnumView} \Pnum{} view}
\end{subfigure}\hfill
\begin{subfigure}[b]{0.47\textwidth}
\includegraphics[height=115pt]{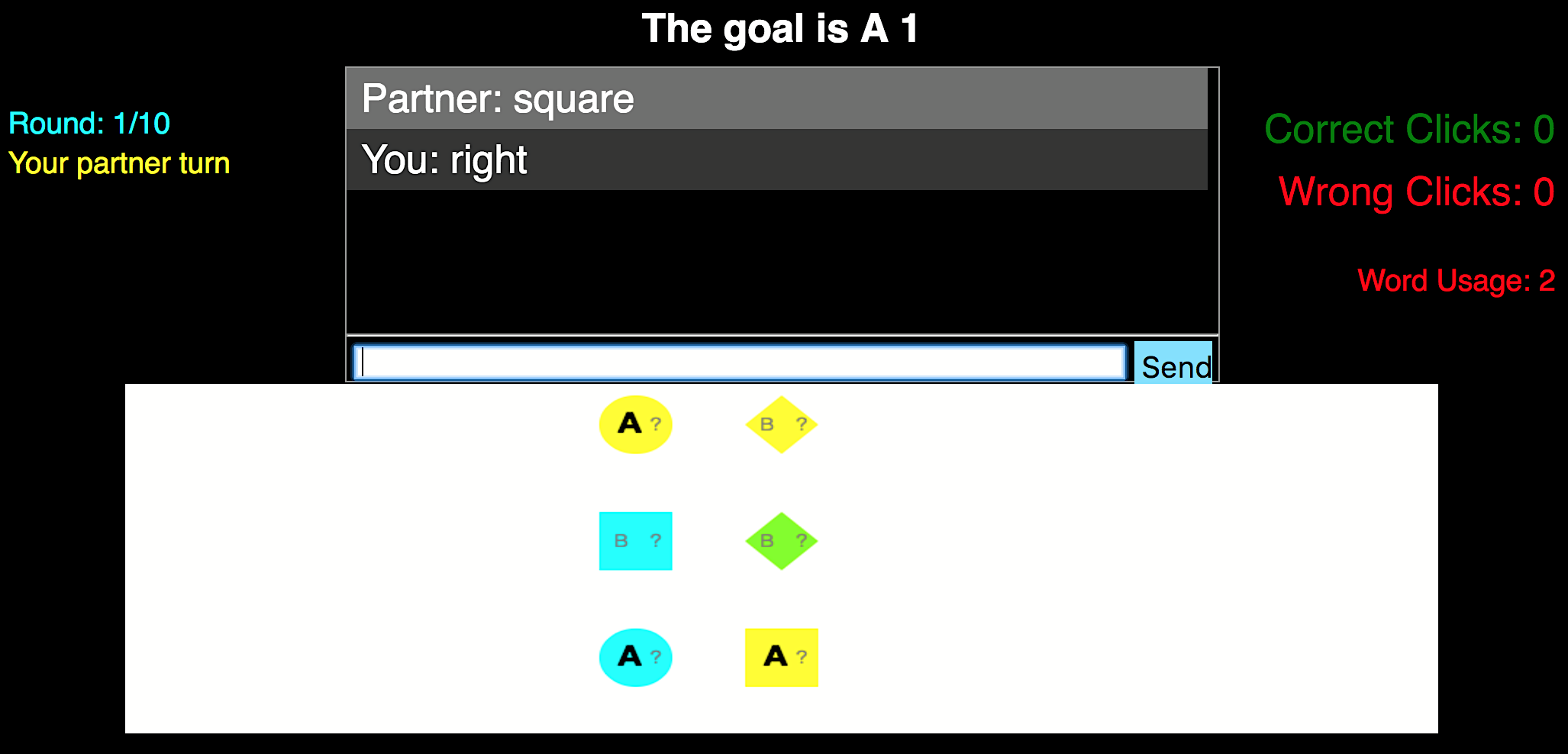}
\caption{\label{fig:pcharitView} \Pchar{} view}
\end{subfigure}
\caption{Chat interface that Amazon Mechanical Turk (AMT) workers use to play \gameName{} (for readability, objects with the goal digit/letter are bolded).}
\label{fig:exampleOfExperiment}
\end{figure*}

In \gameName{} (see \reffig{inference} for an example),
two players try to identify a goal object,
but each only has partial information about its identity.
Thus, in order to solve the task,
they must communicate, piecing their information together like a jigsaw puzzle.
Figure \ref{fig:exampleOfExperiment} shows the interface that humans use to play the game.

More formally,
the shared world state $w$ includes the public properties of a set of objects:
position on a $m \times n$ grid,
color (blue, yellow, green), and
shape (square, diamond, circle).
In addition, $w$ contains the letter and digit of the goal object (e.g., B2).
The private state of player $\Pnum{}$ is a digit (e.g., 1,2,3) for each object,
and the private state of player \Pchar{} is a letter (e.g., A,B,C) for each object.
These states are $s_{+1}, s_{-1}$ depending on which player goes first.

On each turn $t$, a player $j(t)$'s action $a_t$
can be either
(i) a message containing one or two English words \footnote{
If the words are not inside the English dictionary, the sender receives an error and the message is rejected.
This prevents players from circumventing the game rules by connecting multiple words without spaces.
}
(e.g., \spa{circle}),
or (ii) a click on an object, specified by its position (e.g., (1,3)).
A click action terminates the game.
If the clicked object is the goal,
a green square will appear around it which is visible to both players; if the clicked object is not the goal, a red square appears instead.
To discourage random guessing, we prevent players from clicking in the first time step.
Players do not see an explicit utility ($U$); however, they are instructed to think strategically to choose messages that lead to clicking on the correct object while using a minimum number of messages.
Players can see the number of correct clicks, wrong clicks, and number of the words they have sent to each other so far at the top right of the screen.

We would like to study how context-dependent meaning arises out of the interplay between a
context-independent literal semantics with context-sensitive planning, inference, and pragmatics.
The simplicity of the \gameName{} game ensures that this interplay is not obscured by other challenges.

\subsection{Data collection}
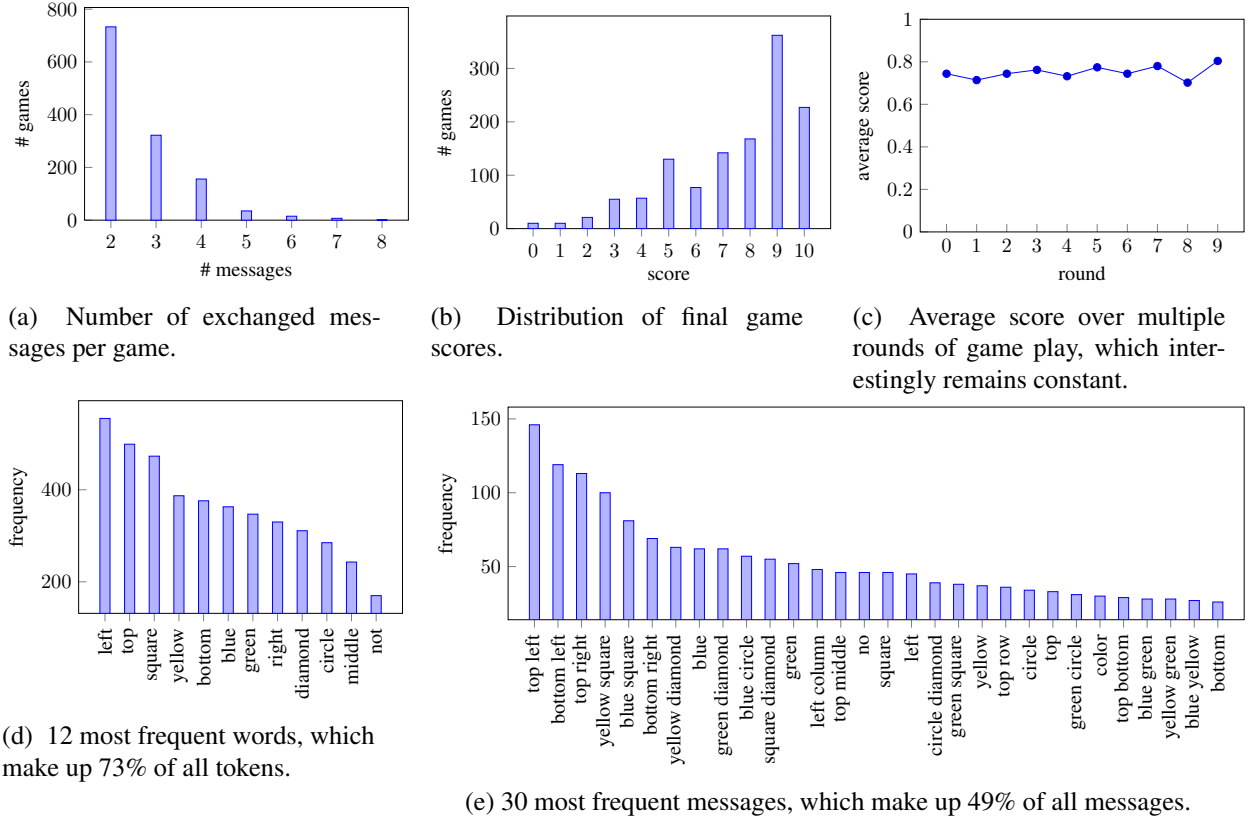
\begin{figure*}[ht]
\pgfplotsset{
	barChartType/.style={enlarge x limits={abs=0.5cm},
	bar width=0.5em,
xtick pos=left,
ytick pos=left,
height=160pt,
width=220pt,
xtick=data,
},
}
\def\sc{0.7}
\centering
 \begin{subfigure}[t]{0.3\textwidth}
\begin{tikzpicture}[scale=\sc]
\begin{axis} [
	ylabel=\# games,
	xlabel=\# messages,ybar,
	barChartType,ybar,ymin=0,
]
\addplot 
	coordinates {	
(2,733) 
  (3,322)
  (4,156)  
  (5,35) 
  (6,15)
  (7,7)
  (8,1)};

\end{axis}
\end{tikzpicture}
   \caption{\label{fig:gamesLength} Number of exchanged messages per game.}
\end{subfigure}\hspace{0.04\textwidth}%
\begin{subfigure}[t]{0.3\textwidth}
\begin{tikzpicture}[scale=\sc]
\begin{axis}[
	ylabel=\# games,
	xlabel=score,ybar,
	barChartType,
    ymin=0,
]

\addplot
	coordinates {
(0,10)
(1,10)
(2,21)
(3,55)
(4,57)
(5,130)
(6,77)
(7,142)
(8,168)
(9,362)
(10,227)
};

\end{axis}
\end{tikzpicture}
  \caption{\label{fig:gamesScore} Distribution of final game scores.}
\end{subfigure}\hspace{0.04\textwidth}%
\begin{subfigure}[t]{0.3\textwidth}
\begin{tikzpicture}[scale=\sc]
  \begin{axis}[
	ymin=0,ymax=1,
    xlabel=round,
	ylabel=average score,
	barChartType,
]
    \addplot coordinates {
(0,.744)
(1,.714)
(2,.744)
(3,.762)
(4,.732)
(5,.774)
(6,.744)
(7,.78)
(8,.702)
(9,.804)
    };
  \end{axis}
\end{tikzpicture}
\caption{\label{fig:gamesScoreOverTime} Average score over multiple rounds of game play, which interestingly remains constant.}
\end{subfigure}

\begin{subfigure}[t]{0.3\textwidth}	

\begin{tikzpicture}[scale=\sc]
\begin{axis}[
barChartType,
xtick=data,
x tick label style={rotate=90,anchor=east},
ybar ,
bar width=0.5em,
	ylabel=frequency,
    symbolic x coords={left,
top,
square,
yellow,
bottom,
blue,
green,
right,
diamond,
circle,
middle,
not,
}
]
\addplot  coordinates{
(left,	555)
(top,	499)
(square,	473)
(yellow,	387)
(bottom,	376)
(blue,	363)
(green,	347)
(right,	330)
(diamond,	311)
(circle,	285)
(middle,	243)
(not,	170)
};
\end{axis}
\end{tikzpicture}
  \caption{\label{fig:wordUsage} 12 most frequent words, which make up 73\% of all tokens. }
\end{subfigure}\hspace{0.04\textwidth}%
\begin{subfigure}[t]{0.65\textwidth}
\raisebox{-25pt}{
\begin{tikzpicture}[scale=\sc]
\begin{axis}[
xtick=data,
x tick label style={rotate=90,anchor=east},
ybar ,
bar width=0.5em,
barChartType,
	ylabel=frequency,
width=1.45*\textwidth,
symbolic x coords={top left ,
bottom left ,
top right ,
yellow square ,
blue square ,
bottom right ,
yellow diamond ,
blue ,
green diamond ,
blue circle ,
square diamond ,
green ,
left column ,
top middle ,
no ,
square ,
left ,
circle diamond ,
green square ,
yellow ,
top row ,
circle ,
top ,
green circle ,
color ,
top bottom ,
blue green ,
yellow green ,
blue yellow ,
bottom ,
}
]

\addplot  coordinates{
(top left ,146)
(bottom left ,119)
(top right ,113)
(yellow square ,100)
(blue square ,81)
(bottom right ,69)
(yellow diamond ,63)
(blue ,62)
(green diamond ,62)
(blue circle ,57)
(square diamond ,55)
(green ,52)
(left column ,48)
(top middle ,46)
(no ,46)
(square ,46)
(left ,45)
(circle diamond ,39)
(green square ,38)
(yellow ,37)
(top row ,36)
(circle ,34)
(top ,33)
(green circle ,31)
(color ,30)
(top bottom ,29)
(blue green ,28)
(yellow green ,28)
(blue yellow ,27)
(bottom ,26)
};
\end{axis}
\end{tikzpicture}}
\caption{\label{fig:messageUsage}30 most frequent messages, which make up 49\% of all messages.}
\end{subfigure}
\caption{\label{fig:datasetStatistics} Statistics of the collected corpus.}

\end{figure*}

We generated 10 \gameName{} scenarios as follows:
For each one, we randomly choose $m, n$ to be either $2 \times 3$ or $3 \times 2$ (which results in $64$ possible private states).
We randomly choose the properties of all objects and randomly designated one as the goal.
We randomly choose either \Pchar{} or \Pnum{} to start the game first.
Finally, to make the scenarios interesting, we keep a scenario if it satisfies:
(i) Only the goal object (and no other objects) has the goal combination of the letter and digit;
(ii) There exist at least two goal-consistent objects for each player and their sum of goal-consistent objects is at least $m \times n$;
and 
(iii) all the goal consistent objects for each player do not share the same color, shape, or position (which means all the goal-consistent objects are not in left, right, top, bottom, or middle).

\begin{figure}[ht]
\centering
\def\sc{0.6}
\def\scc{0.3}
\def\sz{2.8}
\def\di{0.01}
\def\scd{0.7}
\def\sp{0.5cm}
\pgfplotsset{
	barChartType/.style={enlarge x limits={abs=0.5cm},
	bar width=0.5em,
xtick pos=left,
ytick pos=left,
height=160pt,
width=320pt,
xtick=data,
},
}
\begin{tikzpicture}[scale=0.7]
\begin{axis}[
barChartType,
xtick=data,
x tick label style={rotate=90,anchor=east},
ybar ,
bar width=0.5em,
	ylabel=frequency,
    symbolic x coords={
blue square,
middle row,
square diamond,
bottom right,
not top,
color,
middle,
yellow square,
middle left,
square,
not circle,
squares,
middle two,
what color,
yellow circle,
diamond square,
either middle,
maybe middle,
mid row,
middle objects,
}
]
\addplot  coordinates{
(blue square, 19)
(middle row, 13)
(square diamond, 11)
(bottom right, 9)
(not top, 9)
(color, 7)
(middle, 7)
(yellow square, 7)
(middle left, 6)
(square, 6)
(not circle, 4)
(squares, 4)
(middle two, 3)
(what color, 3)
(yellow circle, 3)
(diamond square, 2)
(either middle, 2)
(maybe middle, 2)
(mid row, 2)
(middle objects, 2)
};
\end{axis}
\end{tikzpicture}
\begin{tikzpicture}[scale=\scc]
			\node[goal]  at (0.5*\sz,3*\sz) {Find A1};
			\node[goal]  at (3.5*\sz,3*\sz) {Find A1};
			
			\node[objectC,fill=cyan!20] at (0,0) 		{A ?};
			\node[objectS,fill=yellow!50] at (\sz,0) 		{A ?};
			\node[objectS,fill=cyan!20] at (0,\sz) 		{B ?};
			\node[objectD,fill=green!50] at (\sz,\sz) 	{B ?};
			\node[objectC,fill=yellow!50] at (0,2*\sz) 	{A ?};
			\node[objectD,fill=yellow!50] at (\sz,2*\sz) 	{B ?};

			\node at (0.5*\sz,-\sz) {\small \Pchar{} view };
			
			\draw[dashed] (2*\sz,-\sz) -- (2*\sz,3*\sz);
			
			\node[objectC,fill=cyan!20] at (3*\sz,0) 	{? 3};
			\node[objectS,fill=yellow!50] at (4*\sz,0) 		{? 1};
			\node[objectS,fill=cyan!20] at (3*\sz,\sz) 	{? 1};
			\node[objectD,fill=green!50] at (4*\sz,\sz) 	{? 1};
			\node[objectC,fill=yellow!50] at (3*\sz,2*\sz) {? 2};
			\node[objectD,fill=yellow!50] at (4*\sz,2*\sz) 	{? 2};

			\node at (3.5*\sz,-\sz) {\small \Pnum{} view };
			
			\draw (-0.5*\sz,-1.5*\sz) -- (4.5*\sz,-1.5*\sz);

			\draw (-0.5*\sz,-1.5*\sz) -- (4.5*\sz,-1.5*\sz);
			\node[message] at (-0.5*\sz,-2*\sz)	{\Pnum: \spa{middle}};
			\node[message] at (-0.5*\sz,-2.7*\sz)	{\Pchar: \spa{yellow circle}};
			\node[message] at (-0.5*\sz,-3.4*\sz)	{\Pnum: \spa{bottom right}};
			\node[message] at (-0.5*\sz,-4.1*\sz)	{\Pchar: \spa{click (1,2)}};
			
			\draw (-0.5*\sz,-4.5*\sz) -- (4.6*\sz,-4.6*\sz);
			
			\end{tikzpicture}

    \caption{\label{fig:datasetExample} Bottom: one of the games played by Turkers. Top: the distribution of utterances on the first message.
Players choose to explain their private state in different ways.
Some use more general messages (e.g., \spa{square diamond}),
while some use more specific ones (e.g., \spa{blue square}).
Top diagram shows the first 20 most frequent messages on the first round (72\% of all the messages). 
}
\end{figure}
\begin{table}
\scalebox{0.9}{
\begin{tabular}{lccc}
\toprule
& \bf \# games & \bf \# messages & \bf average score \\
\midrule
\bf All    & 1680 & 4967 & 7.50 \\
\bf Kept   & 1259 & 3358 & 7.48 \\
\bottomrule
\end{tabular}}

\caption{\label{tab:filteredUnfiltered} Statistics for all 1680 games and the 1259 games
in which each message contains at least one of the 12 most frequent words or ``yes'', or ``no''.
}
\end{table}

We collected a dataset of \gameName{} games on Amazon Mechanical Turk using the framework in \citet{hawkins2015conducting} as follows:
200 pairs of players each played all 10 scenarios in a random order.
Out of 200 pairs, 32 pairs left the game prematurely which results in 168 pairs playing the total of 1680 games.
Players performed 4967 actions (messages and clicks) total and obtained an average score (correct clicks) of 7.5 per game.
The average score per scenario varied from 6.4 to 8.2.
Interestingly, there is no significant difference in scores across the 10 scenarios, suggesting that players do not adapt and become more proficient with more game play (\reffig{gamesScoreOverTime}).
\reffig{datasetStatistics} shows the statistics of the collected corpus.
\reffig{datasetExample} shows one of the games,
along with the distribution of messages in the first time step of all games played on this scenario.

To focus on the strategic aspects of \gameName{}, we filtered the dataset to reduce the words in the tail.
Specifically, we keep a game if all its messages
contain at least one of the 12 most frequent words (shown in
\reffig{wordUsage}) or ``yes'' or ``no''.
For example, in \reffig{datasetExample}, the games containing messages such as \spa{what color}, \spa{mid row}, \spa{color} are filtered because they don't contain any frequent words.
Messages such as \spa{middle}, \spa{either middle}, \spa{middle maybe}, \spa{middle objects} are mapped to \spa{middle}.
1259 of 1680 games survived.
\reftab{filteredUnfiltered} compares the statistics between all games and the ones that were kept.
Most games that were filtered out contained less frequent synonyms (e.g. \spa{round} instead of \spa{circle}). Some questions were filtered out too (e.g., \spa{what color}).
Filtered games are 1.15 times longer on average.

\section{Literal Semantics}
\label{sec:semantics}
\begin{figure}
\centering
\def\sc{0.45}
\def\scc{0.25}
\def\sz{2.8}
\def\di{0.01}
\tikzset{
matrixStyle/.style={matrix of math nodes,nodes={scale=0.65, 
minimum height=0.6cm, 
                      anchor=center, align=center,
                      inner sep=0pt, outer sep=0pt},left delimiter={[},right delimiter={]}, inner sep=0,outer sep=0pt},
cross/.style={cross out, draw=black, minimum size=2*(#1-\pgflinewidth), inner sep=0pt, outer sep=0pt},
}

 \begin{subfigure}[b]{0.4\textwidth}
     \centering
     \resizebox{\linewidth}{!}{
       \begin{tikzpicture}[scale=\scc]
\node (s2) at (-0.5*\sz,2.2*\sz) {$s_{-1}$};
\matrix[matrixStyle]  (a5) at (-0.5*\sz,\sz) {
0\\1\\1\\
};
\node (s2) at (4.5*\sz,2.2*\sz) {$s_{+1}$};
\matrix[matrixStyle]  (a5) at (4.5*\sz,\sz) {
1\\0\\1\\
};
			\node[goal,scale=0.8]  at (\sz,3*\sz) {Find B2};
			\node[goal,scale=0.8]  at (3*\sz,3*\sz) {Find B2};
			
			\node[objectC,fill=cyan!20] at (\sz,0) 		{B ?};
			\node[objectS,fill=cyan!20] at (\sz,\sz) 	{B ?};
			\node[objectC,fill=cyan!20] at (\sz,2*\sz) 	{C ?};

			\node[scale=0.8] at (\sz,-\sz) {\small \Pchar{} view };
			
			\draw[dashed] (2*\sz,-\sz) -- (2*\sz,3*\sz);
			
			\node[objectC,fill=cyan!20] at (3*\sz,0) 		{? 2};
			\node[objectS,fill=cyan!20] at (3*\sz,\sz) 	{? 3};
			\node[objectC,fill=cyan!20] at (3*\sz,2*\sz) 	{? 2};

			\node[scale=0.8] at (3*\sz,-\sz) {\small \Pnum{} view };
			
			\end{tikzpicture}
     }
 \end{subfigure}
 \begin{subfigure}[b]{0.4\textwidth}
\begin{align*}
&\den{\text{\spa{square}}} = \left\{s: s\land\bm{0}{1}{0}\neq \bm{0}{0}{0}\right\} \\
&\den{\text{\spa{top bottom}}} = \left\{s: s\land \left(\bm{1}{0}{0} \lor \bm{0}{0}{1} \right)\neq \bm{0}{0}{0}\right\}\\
&\den{\text{\spa{top blue}}} = \left\{s: s\land \left(\bm{1}{0}{0} \land \bm{1}{1}{1} \right)\neq \bm{0}{0}{0}\right\}
\end{align*}
\end{subfigure}

\caption{\label{fig:statesAndActions} Private state of the players and meaning of two action sequences.
}
\end{figure}

In order to understand the principles behind how humans perform planning, inference, and pragmatics, we aim to develop a parsimonious, interpretable model with few parameters
rather than a highly expressive, data-driven model.
Therefore, following the tradition of Rational Speech Acts (RSA) \citep{frank2012pragmatics,goodman2016pragmatic},
we will define in this section a mapping from each word to its \emph{literal semantics},
and rely on the \modelName{} recurrences (which we will describe in \refsec{pip}) to provide context-dependence.
One could also learn the literal semantics by backpropagating through these recurrences,
which has been done for simpler RSA models \citep{monroe2015pragmatics}; or learn the literal semantics from data and then put RSA on top \citep{andreas2016learning};  
we leave this to future work.

Suppose a player utters a single word \spa{circle}.  There are multiple possible context-dependent interpretations:
\begin{itemize}[noitemsep]
\item Are any circles goal-consistent?
\item All the circles are goal-consistent.
\item Some circles but no other objects are goal-consistent.
\item Most of the circles are goal-consistent.
\item At least one circle is goal-consistent.
\end{itemize}
We will show that most of these interpretations can arise from a simple fixed semantics: roughly ``some circle is goal consistent''.
We will now define a simple \emph{literal semantics} of message actions such as \spa{circle},
which forms the base case of \modelName{}.
Recall that the shared world state $w$ contains the goal (e.g., B2) and, assuming $\Pchar{}$ goes first,
the private state $s_{-1}$ ($s_{+1}$)
of player $\Pchar{}$ ($\Pnum{}$) contains the letter (digit) of each object.
For notational simplicity, let us define $s_{-1}$ ($s_{+1}$) to be a matrix corresponding to the spatial locations of the objects,
where an entry is $1$ if the corresponding object has the goal letter (digit) and
$0$ otherwise.
Thus $s_j$ represents also the set of \emph{goal-consistent} objects given the private knowledge of that player.
\reffig{statesAndActions} shows the private states of the
players.

We define two types of message actions: \emph{informative} (e.g., \spa{blue}, \spa{top})
and \emph{verifying} (e.g., \spa{yes}, \spa{no}). 
Informative messages have immediate meaning, while verifying messages depend on the previous utterance.

\paragraph{Informative messages.}

Informative messages describe constraints on the speaker's private state (which the partner does not know).  
For a message $a$, define $\den{a}$ to be the set of consistent private states.
For example, $\den{\spa{bottom}}$ is all private states
where there are goal-consistent objects in the bottom row.

Formally, for each word $x$ that specifies some object property
(e.g., \spa{blue}, \spa{top}),
define $v_x$ to be an $n \times m$ matrix
where an entry is $1$ if the corresponding object has the property $x$, and $0$ otherwise.
Then, define the literal semantics of a single-word message $x$ to be
$\den{x} \eqdef \{ s : s \land v_x \neq 0 \}$, where $\land$ denotes element-wise \emph{and} and $0$ denotes the zero matrix.
That is, single-property messages can be glossed as ``some goal-consistent object has property $x$''.

For a two-word message $xy$,
we define the literal semantics depending on the relationship between $x$ and $y$.
If $x$ and $y$ are mutually exclusive, then we interpret
$xy$ as $x$ \emph{or} $y$ (e.g., \spa{square circle}); otherwise, we interpret
$xy$ as $x$ \emph{and} $y$ (e.g., \spa{blue top}).
Formally,
$\den{xy} \eqdef \{s: s \land (v_x\land v_y) \neq 0\}$ if $v_x \land v_y \neq 0$
and $\{s: s \land (v_x \lor v_y) \neq 0 \}$ otherwise. 
See \reffig{statesAndActions} for some examples.

\paragraph{Action sequences.}

We now define the literal semantics of an entire action sequence
$\den{a_{1:t}}_j$ with respect to player $j$,
which is the set of possible partner private states $s_{-j}$.
Intuitively, we want to simply intersect the set of consistent private states of the informative messages,
but we need to also handle verifying messages (\spa{yes} and \spa{no}),
which are context-dependent.
Formally,
we say that private state $s_{-j} \in \den{a_{1:t}}_j$ if the following hold:
for all informative messages $a_i$ uttered by $-j$,
$s_{-j} \in \den{a_i}$; and for all verifying messages $a_i$ uttered by $-j$
if $a_i = \spa{yes}$ then, $s_{-j} \in \den{a_{i-1}}$; and
if $a_i = \spa{no}$ then, $s_{-j} \not\in \den{a_{i-1}}$.

\section{The Planning-Inference-Pragmatics (\modelName) Model}
\label{sec:pip}
\begin{figure}
\centering
\includegraphics[width=\textwidth/2]{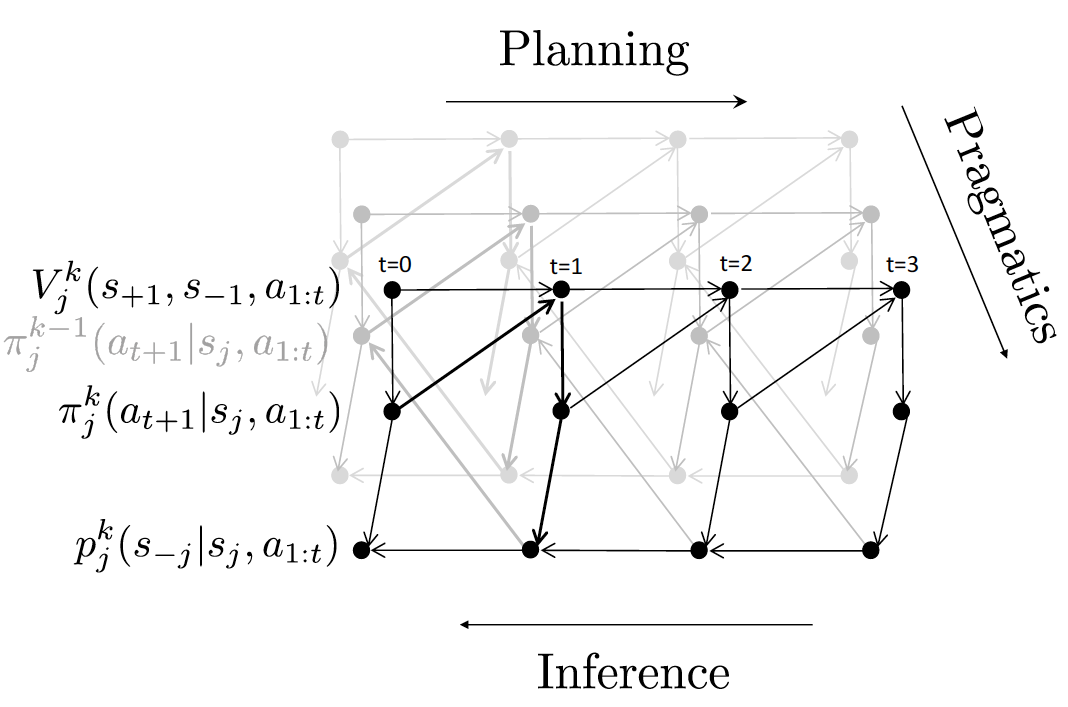}
\caption{
\modelName{} is defined via a system of recurrences
that simultaneously captures planning, inference, and pragmatics.
The arrows show the dependencies between beliefs $p$, expected utilities $V$, and policy $\pi$.}
\label{fig:allInOne}
\end{figure}

Why does \Pnum{} in \reffig{inference} choose \spa{circle} rather than \spa{top} or \spa{click(1,2)}?
Intuitively,
when a player chooses an action, she should take into account her
previous actions, her partner's actions, and the effect of her action on future turns.
She should do all these while reasoning pragmatically that her partner is also a strategic player.

At a high-level, \modelName{} defines a system of recurrences revolving around three concepts,
depicted in \reffig{allInOne}:
player $j$'s beliefs over the partner's private state $p_j^k(s_{-j} \mid s_j, a_{1:t})$,
her expected utility of the game $V_j^k(s_{+1}, s_{-1}, a_{1:t})$, and
her policy $\pi_j^k(a_t \mid s_j, a_{1:t-1})$.
Here, $t$ indexes the current time and $k$ indexes the depth of pragmatic recursion,
which will be explained later in \refsec{pragmatics}.
To simplify the notation, we have dropped $w$ (shared world state) from the
notation, since everything conditions on it.

\subsection{Inference}

\label{sec:inference}
From player $j$'s point of view,
the purpose of inference is to compute a distribution over the partner's private state $s_{-j}$
given all actions thus far $a_{1:t}$.
We first consider a ``level 0'' player,
which simply assigns a uniform distribution
over all states consistent with the literal semantics of $a_{1:t}$,
which we defined in \refsec{semantics}:
\begin{align}
p_j^0(s_{-j} \mid s_j, a_{1:t}) &\propto \begin{cases}
  1\quad \text{$s_{-j} \in \den{a_{1:t}}_j$}, \\
  0\quad \text{otherwise}.
\end{cases}
\label{eqn:inference}
\end{align}

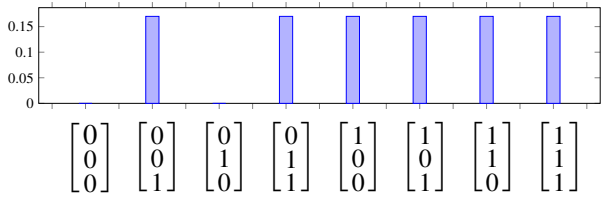
\begin{figure}[h]
\centering
\def\sc{0.9}
\def\dis{0.99}
\def\diss{2}
\tikzset{
matrixStyle/.style={matrix of math nodes,nodes={scale=\sc, 
                      anchor=center, align=center,
                      inner sep=0pt, outer sep=0pt},left delimiter={[},right delimiter={]}, inner sep=0,outer sep=0pt, row sep=0.2em,},
}
\begin{tikzpicture}[scale=0.5]
\begin{axis}[
	ybar,
	height=\textwidth/4,width=\textwidth,
xticklabels={},ymin=0,
ytick={0,0.05,0.1,0.15,0.2},
yticklabels={0,0.05,0.1,0.15,0.2},
]
\addplot
	coordinates {
(0,0)
(1,0.17)
(2,0)
(3,0.17)
(4,0.17)
(5,0.17)
(6,0.17)
(7,0.17)
};

\end{axis}
\end{tikzpicture}
\begin{tikzpicture}[scale=0.9]
\node at (-0.9,0) {};
\matrix[matrixStyle] (a0) at (0,0) {
\large 0\\0\\0\\
};

\matrix[matrixStyle] (a1) at (\dis,0) {
0\\0\\1\\
};
\matrix[matrixStyle] (a2) at (\dis*2,0) {
0\\1\\0\\
};
\matrix[matrixStyle] (a3) at (\dis*3,0) {
0\\1\\1\\
};
\matrix[matrixStyle] (a4) at (\dis*4,0) {
1\\0\\0\\
};
\matrix[matrixStyle] (a5) at (\dis*5,0) {
1\\0\\1\\
};
\matrix[matrixStyle] (a6) at (\dis*6,0) {
1\\1\\0\\
};
\matrix[matrixStyle] (a7) at (\dis*7,0) {
1\\1\\1\\
};


\end{tikzpicture}
\caption{\label{fig:literalInference} \Pchar{}'s probability distribution over
\Pnum{}'s private state after \Pnum{} says \spa{circle} in the game shown in
\reffig{statesAndActions}.
}
\end{figure}

For example, \reffig{literalInference}, shows the $\Pchar{}$'s belief about $\Pnum{}$'s private state after observing \spa{circle}.
Remember we show the private state of the players as a matrix where an entry is $1$ if the corresponding object has the goal letter (digit) and
$0$ otherwise. 

A player's own private state $s_j$ can also constrain her beliefs
about her partner's private state $s_{-j}$.
For example, in \gameName{},
the active player knows there is a goal,
and so we set
$p_j^k(s_{-j} \mid s_j,a_{1:t}) = 0$ if $s_{-j} \land s_j = 0$.
\subsection{Planning}

\label{sec:planning}
The purpose of planning is to compute a policy $\pi_j^k$,
which specifies a distribution over
player $j$'s actions $a_t$ given all past actions $a_{1:t-1}$.
To construct the policy, we first define an expected utility $V_j^k$
via the following forward-looking recurrence:
When the game is over (e.g., in \gameName{}, one player clicks on an object),
the expected utility of the dialogue is simply its utility as defined by the game:
\begin{align}
  V_j^k(s_{+1},s_{-1},a_{1:t}) &= U(s_{+1}, s_{-1}, a_{1:t}).
\label{eqn:utility1}
\end{align}
Otherwise, we compute the expected utility assuming that in the next turn,
player $j$ chooses action $a_{t+1}$ with probability governed by her policy
$\pi_{j}^k(a_{t+1} \mid s_{j}, a_{1:t})$:
\begin{align} 
  \nonumber  V_j^k(s_{+1},s_{-1},a_{1:t}) =
 \sum_{a_{t+1}} &\pi_j^k(a_{t+1} \mid s_j, a_{1:t})\\
  &V_{-j}^k(s_{-1}, s_{+1}, a_{1:t+1}).
\label{eqn:expectedUtility}
\end{align}

\begin{figure}[ht]
\centering
\includegraphics[width=\textwidth/2]{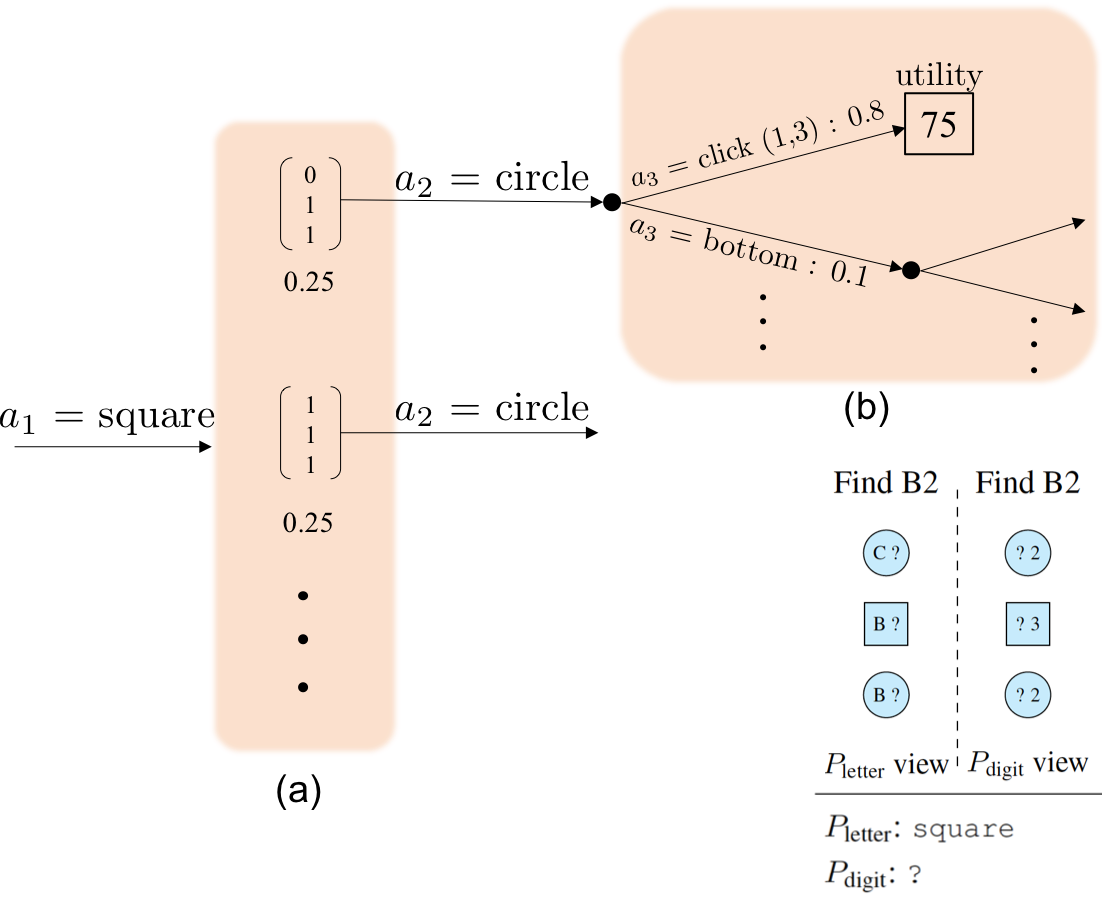}
\caption{\label{fig:planning}
Planning reasoning for the game in Figure 1 (reproduced here in the bottom right). (a) In order to calculate the expected gain ($E$) of generating \spa{circle},
for every state $s$, \Pnum{} computes the probability of $s$ being the \Pchar{}'s private state.
(b) She then computes the expected utility ($V$) if she generates \spa{circle} assuming \Pchar{}'s private state is $s$. 
}
\end{figure}

Having defined the expected utility, we now define the policy.
First, let $D_j^k$ be the gain in expected utility $V_{-j}^k(s_{+1},s_{-1},a_{1:t})$
over a simple baseline policy that ends
the game immediately, yielding utility $U(s_{+1},s_{-1},a_{1:t-1})$ (which is simply a penalty for not finding the correct goal and a penalty for each action).
Of course, the partner's private state $s_{-j}$ is unknown and must be marginalized out based on player $j$'s beliefs;
let $E_j^k$ be the expected gain.
Let the probability of an action $a_t$ be proportional to $\max(0,E_j^k)^\alpha$,
where $\alpha \in [0, \infty)$ is a hyperparameter that controls the rationality of the agent
(a larger $\alpha$ means that the player chooses utility-maximizing actions more aggressively).
Formally:
\begin{align}
  &D_j^k = V_{-j}^k(s_{+1}, s_{-1}, a_{1:t}) - U(s_{+1},s_{-1},a_{1:t-1}), \nonumber \\
  &E_j^k = \sum_{s_{-j}} p_j^k(s_{-j} \mid s_j, a_{1:t-1}) D_j^k, \nonumber \\
  &\pi_j^k(a_{t} \mid s_j, a_{1:t-1}) \propto \max\left(0, E_j^k\right)^\alpha.
\label{eqn:generating}
\end{align}
In practice, we use a depth-limited recurrence,
where the expected utility is computed assuming that the game will end in $f$ turns and the last action is a click action (meaning that we only consider the action sequences with size $\leq f$ and a clicking action as the last action).
\reffig{planning} shows how \Pnum{} computes the expected gain \refeqn{generating} of saying \spa{circle}.

\begin{figure*}[t]
\centering
\includegraphics[width=\textwidth]{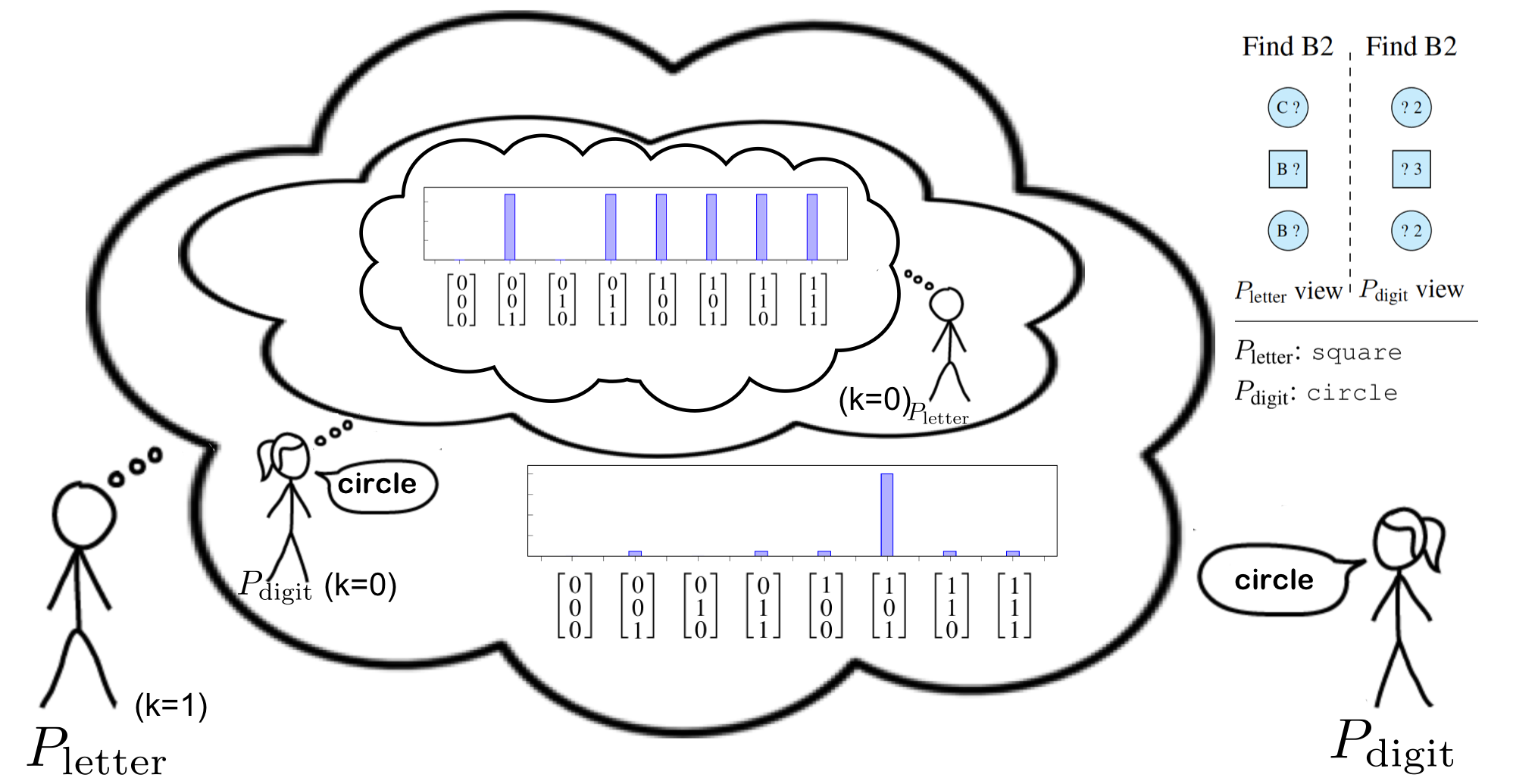}
\caption{\label{fig:pragmatics}
Pragmatic reasoning for the game in Figure 1 (reproduced here in the upper right) at time step 3.
Players reason recursively about each others beliefs:
the level-$0$ player puts a uniform distribution $p_j^0$ over all the states in which at least one circle is goal-consistent independent of the shared world state and previous actions.
The level-$1$ player assigns probability over the partner's private states $s_{-j}$ proportional to the probability that she would have performed the last action given that state $s_{-j}$.
For example, if $\protect\bm{0}{0}{1}$ were \protect\Pnum{}'s private state, then saying \spa{bottom} would be more probable (given the shared world state); 
if $\protect\bm{1}{1}{1}$ were \protect\Pnum{}'s state, then clicking on the square would be a better option (given the previous actions).
But given that \protect\Pnum{} uttered \protect\spa{circle}, $\protect\bm{1}{0}{1}$ is most likely, as reflected by $p_j^1$.
}
\end{figure*}

\subsection{Pragmatics}
\label{sec:pragmatics}

The purpose of pragmatics is to take into account the partner's strategizing.
We do this by constructing a level-$k$ player that infers the partner's private state,
following the tradition of Rational Speech Acts (RSA) \cite{frank2012pragmatics,goodman2016pragmatic}.
Recall that a level-0 player $p_j^0$ (\refsec{inference}) puts a uniform distribution over all the semantically valid private states of the partner.
A level-$k$ player assigns probability over the partner's private state
proportional to the probability that a level-$(k-1)$ player would have performed the last action $a_t$:
\begin{align}
  \nonumber p^k_j(s_{-j} \mid s_j, a_{1:t}) &\propto  \pi^{k-1}_{-j} (a_t \mid s_{-j}, a_{1:t-1})
  \\ & \quad\quad\quad p^{k}_j(s_{-j} \mid s_j, a_{1:t-2}).
\label{eqn:pragmatics}
\end{align}

\reffig{pragmatics} shows an example of the pragmatic reasoning.

\subsection{A closer look at the meaning of actions}
\label{sec:richerMeanings}


In the \refsec{planning}, we modeled the players as rational agents that choose actions that lead to higher gain utility.
In the pragmatics section (\refsec{pragmatics}), we described how a player infers the partner's private state taking into account that her partner is acting cooperatively.
The phenomena that emerges from the combination of the two is the topic of this section.

We first define the belief marginals $B_j$ of a player $j$ to be the marginal
probabilities that each object is goal-consistent under the hypothesized partner's private state $s_{-j} \in \R^{m \times n}$,
conditioned on actions $a_{1:t}$:
\begin{align}
  B_j(s_j, a_{1:t})&= \sum_{s_{-j}} p_j^k(s_{-j} \mid s_j, a_{1:t}) s_{-j}. 
\label{eqn:beliefState}
\end{align}
At time $t=0$ (before any actions),
the belief marginals of both players are $m \times n$ matrices with $0.5$ in all entries.
The change in a belief marginal after observing an action $a_t$ gives a sense of the effective (context-dependent) meaning of that action. 

We first explain how pragmatics ($k>0$ in \refeqn{pragmatics}) leads to rich action meanings.
When a player observes her partner's action $a_t$,
she assumes this action was chosen because it results in a higher utility than
the alternatives.
In other words, she infers that her partner's private state cannot be one
in which $a_t$ does not lead to high utility.
As an example, saying \spa{circle} instead of \spa{top circle} or \spa{bottom circle} implies that there is more than one goal-consistent circle.
The pragmatic depth $k$ governs the extent to which this type of reasoning
is applied.

Recall in \refsec{planning},
a player chooses an action conditioned on all previous actions,
and the other player assumed this context-dependence.
As an example, \reffig{beliefStates}(d) shows how \spa{right} changes its meaning when it follows \spa{bottom}.

\input{internalStateFig}

\section{Experiments}

\subsection{Setup}
We \emph{a priori} set the reward of clicking on the goal to be $+100$ and clicking on the wrong object to be $-100$.
We set the smoothing $\alpha=10$ and the action cost to be $-50$ based on the data.
The larger the action cost, the fewer messages will be used before selecting an object.
Formally, after $k$ actions:
\begin{align}
  \text{Utility} = -50 k + \begin{cases}
  +100\quad \text{\small the goal object is clicked}, \\
  -100\quad \text{\small otherwise}.
\end{cases}
\label{eqn:utility}
\end{align}

We smoothed all polices by adding $0.01$ to the probability of each action and re-normalizing.
By default, we set $k=1$ (pragmatic depth \refeqn{generating}).
When computing the expected utility \refeqn{expectedUtility} of the game, we use a lookahead of $f=2$.
Inference looks back $b$ time steps (i.e.~\refeqn{inference} and \refeqn{pragmatics} are based on $a_{t-b+1:t}$ rather than $a_{1:t}$);
we set $b=\infty$ by default.

We implemented two baseline policies:

\noindent{\bf Random policy:} for player $j$, the random policy randomly chooses one of the semantically valid (\refsec{semantics}) actions with respect to $s_j$
or clicks on a goal-consistent object.
Formally, the random policy places a uniform distribution over:
\begin{align}
  \{a : s_j \in \den{a}\} \cup \{\text{\spa{click}}(u,v): (s_j)_{u,v}=1\}.
\end{align}

\noindent{\bf Greedy Policy:} assigns higher probability to the actions that convey more information about the player's private state. 
We heuristically set the probability of generating an action proportional to how much it shrinks the set of semantically valid states.
Formally, for the message actions:
\begin{align}
  \pi_j^\text{msg}(a_t \mid a_{1:t-1},s_j) \propto |\den{a_{1:t-1}}_{-j}| - |\den{a_{1:t}}_{-j}|
\end{align}
For the clicking actions, we compute the belief state as explained in \refsec{richerMeanings}. 
Remember $B_{u,v}$ is the marginal probability of the object in the row $u$ and column $v$ being goal-consistent in the partner's private state.
Formally, for clicking actions:
\begin{align}
  &\pi_j^\text{click}(\text{\spa{click}}(u, v) \mid a_{1:t},s_j) \propto \nonumber \\ & \quad\quad\quad\quad \min ({(s_j)}_{u,v}, B_j(s_j,a_{1:t})_{u,v}).
\end{align}
Finally, the greedy policy chooses a click action with probability $\gamma$ and a message action with probability $1-\gamma$.
So that $\gamma$ increases as the player gets more confident about the position of the goal,
we set $\gamma$ to be the probability of the most probable position of the goal:
$\gamma = \max\limits_{u,v} \pi_j^\text{click}(\text{\spa{click}}(u, v) \mid a_{1:t}, s_j)$.

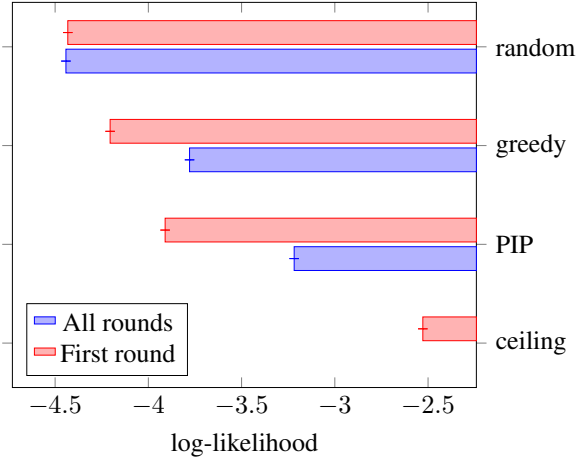
\begin{figure}[t]
\def\dc{0.9}
\pgfplotsset{
	myType/.style={enlargelimits=0.15,
xlabel= log-likelihood,
ylabel={},
xbar,
yticklabel pos=right
},
}

\centering
\begin{tikzpicture}[scale=\dc]
\pgfplotsset{
    /pgfplots/xbar legend/.style={
    /pgfplots/legend image code/.code={%
       \draw[##1,/tikz/.cd]
        (0cm,0cm) rectangle (0.8em,3pt);},
   },
}
\begin{axis}[
myType,
yticklabels={ceiling, \modelName{}, greedy, random },
ytick={0,1,2,3},
legend pos=south west,
]

\addplot+ [error bars/.cd,y dir=both, y explicit]
 coordinates {
(-4.442,3)+=(0.0030000000000001137,0)-=(0.0019999999999997797,0)
(-3.779,2)+=(0.02300000000000013,0)-=(0.02400000000000002,0)
(-3.218,1)+=(0.039000000000000146,0)-=(0.03699999999999992,0)
};

\addplot+[error bars/.cd,y dir=both, y explicit]
 coordinates {
(-4.432,3)+=(0.002000000000000668,0)-=(9.999999999994458E-4,0)
(-4.205,2)+=(0.022999999999999687,0)-=(0.02200000000000024,0)
(-3.91,1)+=(0.04200000000000026,0)-=(0.040999999999999925,0)
(-2.528,0) += (0.04383054,0) -= (0.04773946,0)
};

\legend{All rounds, First round}
\end{axis}
\end{tikzpicture}
\caption{Average log-likelihood across messages.
(a) Performance of \modelName{} and baselines on all time steps.
(b) Performance of \modelName{} and baselines on only the first time step
  along with the ceiling given by the entropy of the human data.
  The error bars show 90\% confidence intervals.
\label{fig:baselineComparisonLogLikelihood}
  }
\end{figure}
\begin{figure*}
\usepgfplotslibrary{statistics}
\def\dc{0.64}
\centering
\begin{subfigure}[b]{0.32\textwidth}
\begin{tikzpicture}[scale=\dc]
\pgfplotsset{
 boxplot width/.initial=0.1em,
}
\begin{axis}[
boxplot/draw direction=y,
 xticklabels={\modelName{}, \noPrag, \noPlan, \noInfer},xtick={1,2,3,4},
  x tick label style={rotate=90,anchor=east},
ylabel=log-likelihood,
]


\addplot+ [boxplot prepared={
draw position=1,
lower whisker=-3.255, 
lower quartile=-3.238,
median=-3.218, 
upper quartile=-3.199,
upper whisker=-3.179,
box extend=0.3,
},
] coordinates {};
\addplot+ [boxplot prepared={
draw position=2,
lower whisker=-3.475, 
lower quartile=-3.459,
median=-3.442, 
upper quartile=-3.425,
upper whisker=-3.41,
box extend=0.3,
},
] coordinates {};
\addplot+ [boxplot prepared={
draw position=3,
lower whisker=-3.313, 
lower quartile=-3.297,
median=-3.277, 
upper quartile=-3.254,
upper whisker=-3.237,
box extend=0.3,
},
] coordinates {};
\addplot+ [boxplot prepared={
draw position=4,
lower whisker=-3.245, 
lower quartile=-3.226,
median=-3.205, 
upper quartile=-3.185,
upper whisker=-3.164,
box extend=0.3,
},
] coordinates {};
\end{axis}
\end{tikzpicture}
\caption{Performance over all games and all rounds.}
\end{subfigure}\hspace{0.04\textwidth}%
\begin{subfigure}[b]{0.32\textwidth}
\begin{tikzpicture}[scale=\dc]
\begin{axis}[
boxplot/draw direction=y,
 xticklabels={\modelName{}, \noPrag, \noPlan, \noInfer},xtick={1,2,3,4},
  x tick label style={rotate=90,anchor=east},
ylabel=log-likelihood,
]

\addplot+ [boxplot prepared={
draw position=1,
lower whisker=-2.482, 
lower quartile=-2.415,
median=-2.331, 
upper quartile=-2.258,
upper whisker=-2.19,
box extend=0.3,
},
] coordinates {};
\addplot+ [boxplot prepared={
draw position=2,
lower whisker=-2.67, 
lower quartile=-2.605,
median=-2.528, 
upper quartile=-2.459,
upper whisker=-2.393,
box extend=0.3,
},
] coordinates {};
\addplot+ [boxplot prepared={
draw position=3,
lower whisker=-2.539, 
lower quartile=-2.467,
median=-2.386, 
upper quartile=-2.31,
upper whisker=-2.242,
box extend=0.3,
},
] coordinates {};
\addplot+ [boxplot prepared={
draw position=4,
lower whisker=-2.63, 
lower quartile=-2.56,
median=-2.477, 
upper quartile=-2.398,
upper whisker=-2.324,
box extend=0.3,
},
] coordinates {};
\end{axis}
\end{tikzpicture}
\caption{Performance over messages after round 3.}
\end{subfigure}\hspace{0.04\textwidth}%
\begin{subfigure}[b]{0.28\textwidth}
\scalebox{0.78}{
\raisebox{7em}{\begin{tabular}{lcccc}
\toprule&\rotatebox{90}{\modelName}&
\rotatebox{90}{\noPrag}&
\rotatebox{90}{\noPlan}&
\rotatebox{90}{\noInfer} \\ \midrule
$k$ \tiny(pragmatics) &	1		& 0	& 1 		& 1\\
$f$ \tiny(planning) &	2		& 2	& 1 		& 2\\
$b$ \tiny(inference) &	$\infty$& $\infty$ & $\infty$	& 1\\ \midrule 

rank all 	&	17.1	& 19.3	& 17.2 & 16.9\\
rank $\ge3$ &	10.4	& 10.8	& 11.6 & 13.1\\
\bottomrule
\end{tabular}}}
\caption{Top: parameter setup. Bottom: expected ranking of human messages according to the different ablations}
\end{subfigure}
\caption{\label{fig:modelsParameters}
  Performance on ablations of PIP.
  Average log-likelihood per message,
  the whiskers show 90\% confidence intervals.
\modelName{} has better performance of ablation of planning and pragmatics over all rounds. 
Looking only one step backward has a better performance in the first few rounds but it is worse after round 3.
}
\end{figure*}
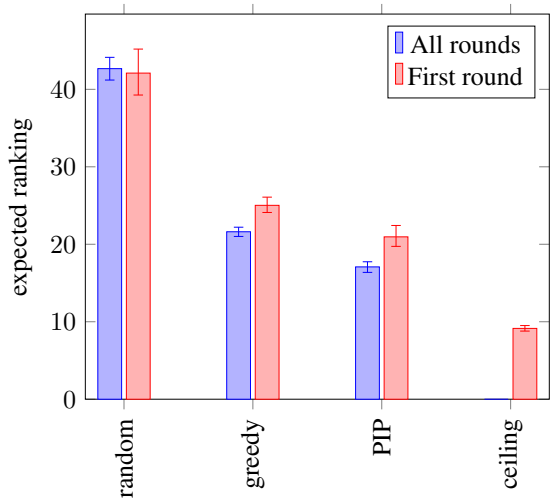
\begin{figure}[t]
\def\dc{0.9}
\centering
\begin{tikzpicture}[scale=\dc]
\pgfplotsset{
    /pgfplots/ybar legend/.style={
    /pgfplots/legend image code/.code={%
       \draw[##1,/tikz/.cd,yshift=-0.25em]
        (0cm,0cm) rectangle (3pt,0.8em);},
   },
}

\begin{axis}[
ylabel=expected ranking, ybar,
  xticklabels={random,greedy,\modelName{}, ceiling},
	xtick={0,1,2,3},
  x tick label style={rotate=90,anchor=east}, 
xtick=data,
legend pos=south west,
ymin=0,
legend pos=north east,
]
\addplot+[error bars/.cd,y dir=both, y explicit]
coordinates {
(0,42.671)+=(0,1.4590000000000032)-=(0,1.472999999999999)
(1,21.615)+=(0,0.588000000000001)-=(0,0.6039999999999992)
(2,17.08)+=(0,0.6620000000000026)-=(0,0.7089999999999996)
(3,0)
};

\addplot+[error bars/.cd,y dir=both, y explicit]
coordinates {
(0,42.095)+=(0,3.1000000000000014)-=(0,2.8299999999999983)
(1,25.026)+=(0,1.0579999999999998)-=(0,0.9209999999999994)
(2,20.958)+=(0,1.4690000000000012)-=(0,1.2289999999999992)
(3,9.14)+=(0,0.3669999999999991)-=(0,0.3470000000000013)
};

\legend{All rounds, First round}
\end{axis}
\end{tikzpicture}
  \caption{\label{fig:ranking} Expected ranking of the  human messages according to different policies.
  Error bars show 90\% confidence intervals.
  }
\end{figure}
\subsection{Results}
\reffig{baselineComparisonLogLikelihood} compares the two baselines with \modelName{} on the
task of predicting human behavior as measured by log-likelihood.\footnote{We
bootstrap the data 1000 times and we show 90\% confidence intervals.}
To estimate the best possible (i.e.~ceiling) performance,
we compute the entropy of the actions on the first time step based on approximately 100 data points per scenario.
For each policy, we rank the actions by their probability in decreasing order
(actions with the same probability are randomly ordered),
and then compute the average ranking across actions according to the different policies; see \reffig{ranking} for the results.

To assess the different components (planning, inference, pragmatics) of \modelName{}, we run \modelName{}, ablating one component at a time
from the default setting of $k=1$, $f=2$, and $b=\infty$ (see \reffig{modelsParameters}).

\paragraph{Pragmatics.}
Let \noPrag{} be \modelName{} but with a pragmatic depth \refeqn{generating} of $k=0$ rather than $k=1$,
which means that \noPrag{} only draws inferences based on the literal semantics of messages.
\noPrag{} loses $0.21$ in average log-likelihood per action, highlighting the importance of pragmatics in modeling human behavior.

\paragraph{Planning.}
Let \noPlan{} be \modelName{}, but looking ahead only $f=1$ step when computing the expected utility \refeqn{expectedUtility} rather than $f=2$.
With a shorter future horizon,
\noPlan{} tries to give as much information as possible at each turn,
whereas human players tend to give information about their state incrementally.
\noPlan{} cannot capture this behavior and allocates low probability to these kinds of dialogue.
\noPlan{} has an average log-likelihood which is $0.05$ lower than that of \modelName{},
highlighting the importance of planning.

\paragraph{Inference.}
Let \noInfer{} be \modelName{}, but only looking at the last utterance ($b=1$) rather than the full history ($b=\infty$).
The results here are more nuanced.
Although \noInfer{} actually performs better than \modelName{} on all games,
we find that \noInfer{} is worse than \modelName{} by an average log-likelihood of $0.15$ in predicting messages after time step 3,
highlighting the importance of inference, but only in long games. 
It is likely that additional noise involved in the inference process leads to the decreased performance when backward looking inference is not actually needed. 
\section{Related Work and Discussion}

Our work touches on ideas in game theory, pragmatic modeling, dialogue modeling, and learning communicative agents,
which we highlight below.

\paragraph{Game theory.}

According to game theory terminology \cite{shoham2008multiagent},
\gameName{} is a non-cooperative (there is no offline optimization of the player's policy before the game starts),
common-payoff (the players have the same utility),
incomplete information (the players have private state)
game with the sequential actions.
One related concept in game theory related to our model is rationalizability
\cite{bernheim1984rationalizable,pearce1984rationalizable}.
A strategy is rationalizable if it is justifiable to play against a completely rational player.
Another related concept is epistemic games \cite{dekel2015epistemic,perea2012epistemic}.
Epistemic game theory studies the behavioral implications of rationality and mutual beliefs in games.

It is important to note that we are not interested in notions of global optima or equilibria;
rather, we are interested in modeling human behavior.
Restricting words to a very restricted natural language
has been studied in the context of language games
\cite{wittgenstein1953philosophical,lewis2008convention,nowak1999evolutionary,franke2009signal,huttegger2010evolutionary}.

\paragraph{Rational speech acts.}
The pragmatic component of \modelName{} is based on Rational
Speech Act framework \citep{frank2012pragmatics,golland2010pragmatics},
which defines recurrences capturing how one agent reasons about another.
Similar ideas were explored in the precursor work of \citet{golland2010pragmatics},
and much work has ensued
\citep{smith2013pragmatics,qing2014gradable,monroe2015pragmatics,ullman2016pragmatics,andreas2016reasoning}. 

Most of this work is restricted to production and comprehension of a single utterance.
\citet{hawkins2015you} extend these ideas to two utterances (a question and an answer).
\citet{vogel2013implicatures} integrates planning with pragmatics using decentralized partially observable Markov processes (DEC-POMDPs).
In their task, two bots should find and co-locate with a specific card. 
In contrast to \gameName{}, their task can be completed without communication; their agents only communicate once sharing the card location. 
They also only study artificial agents playing together
and were not concerned about modeling human behavior.

\paragraph{Learning to communicate.}
There is a rich literature on multi-agent reinforcement learning
\cite{busoniu2008comprehensive}.
Some works assume full visibility and cooperate without communication,
assuming the world is completely visible to all agents
\cite{lauer2000algorithm,littman2001value};
others assume a predefined convention for communication \cite{zhang2013coordinating,tan1993multi}.
There is also some work that learns the convention itself \cite{foerster2016learning,sukhbaatar2016learning,lazaridou2017multi,mordatch2018emergence}.
\citet{lazaridou2017multi} puts humans in the loop to make the communication more human-interpretable.
In comparison to these works,
we seek to predict human behavior instead of modeling artificial agents that
communicate with each other.

\paragraph{Dialogue.}
There is also a lot of work in computational linguistics and NLP on modeling dialogue.
\citet{allen1980analyzing} provides a model that infers the intention/plan of the other agent and uses this plan to generate a response.
\citet{clark1991grounding} explains how two players update their common ground (mutual knowledge, mutual beliefs, and mutual assumptions) in order to coordinate.
Recent work in task-oriented dialogue uses POMDPs and end-to-end neural networks
\citep{young2000probabilistic,young2013pomdp,wen2017network,he2017symmetric}.
In this work, instead of learning from a large corpus, we predict human behavior without learning,
albeit in a much more strategic, stylized setting (two words per utterance).
\section{Conclusion}

In this paper, we started with the observation that humans use language
in a very contextual way driven by their goals.  
We identified three salient aspects---planning, inference, pragmatics---and
proposed a unified model, \modelName{}, that captures all three aspects simultaneously.
Our main result is that a very simple, context-independent literal semantics
can give rise via the recurrences to rich phenomena.
We study these phenomena in a new game, \gameName{},
and show that \modelName{} is able to capture human behavior.

\section*{Reproducibility} 
All code, data, and
experiments for this paper are available
on the CodaLab platform at \url{https://worksheets.codalab.org/worksheets/0x052129c7afa9498481185b553d23f0f9/}.

\section*{Acknowledgments}
We would like to thank the anonymous reviewers and the action editor for their helpful comments.
We also thank Will Monroe for providing valuable feedback on early drafts.

\bibliographystyle{acl2012}
\bibliography{refdb/all}

\end{document}